\documentclass[]{interact}

\usepackage{epstopdf}
\usepackage[caption=false]{subfig}

\usepackage[numbers,sort&compress]{natbib}
\bibpunct[, ]{[}{]}{,}{n}{,}{,}
\renewcommand\bibfont{\fontsize{10}{12}\selectfont}
\makeatletter
\def\NAT@def@citea{\def\@citea{\NAT@separator}}
\makeatother

\usepackage{multirow}

\theoremstyle{plain}

\theoremstyle{definition}

\theoremstyle{remark}

\usepackage[normalem]{ulem}

\usepackage{adjustbox}
\usepackage[table,xcdraw]{xcolor}
\usepackage{pdfcomment}

\usepackage{fancyhdr}
\fancypagestyle{firstpage}{%
  \lhead{To appear in Advanced Robotics, Vol. 38, Issue 21 (Nov 2024)}
  \rhead{}
}

\begin{document}

\articletype{ARTICLE TEMPLATE}

\title{A Cranial-Feature-Based Registration Scheme for Robotic Micromanipulation Using a Microscopic Stereo Camera System}

\author{
\name{Xiaofeng Lin\textsuperscript{a}, Saúl Alexis Heredia Pérez\textsuperscript{b} and Kanako Harada\textsuperscript{b}\thanks{CONTACT Kanako Harada. Email: kanakoharada@g.ecc.u-tokyo.ac.jp}}
\affil{\textsuperscript{a}Department of Bioengineering, The University of Tokyo, Japan;\\ \textsuperscript{b}Graduate School of Medicine, The University of Tokyo, Japan; }
}

\markboth{To appear in Advanced Robotics Vol. 00, No. 00, November 2024, 1–18}
{Shell \MakeLowercase{\textit{et al.}}: A Sample Article Using IEEEtran.cls for IEEE Journals}%

\maketitle

\thispagestyle{firstpage}

\begin{abstract}
Biological specimens exhibit significant variations in size and shape, challenging autonomous robotic manipulation. We focus on the mouse skull window creation task to illustrate these challenges. The study introduces a microscopic stereo camera system (MSCS) enhanced by the linear model for depth perception. Alongside this, a precise registration scheme is developed for the partially exposed mouse cranial surface, employing a CNN-based constrained and colorized registration strategy. These methods are integrated with the MSCS for robotic micromanipulation tasks. The MSCS demonstrated a high precision of 0.10 mm ± 0.02 mm measured in a step height experiment and real-time performance of 30 FPS in 3D reconstruction. The registration scheme proved its precision, with a translational error of 1.13 mm ± 0.31 mm and a rotational error of 3.38° ± 0.89° tested on 105 continuous frames with an average speed of 1.60 FPS. This study presents the application of a MSCS and a novel registration scheme in enhancing the precision and accuracy of robotic micromanipulation in scientific and surgical settings. The innovations presented here offer automation methodology in handling the challenges of microscopic manipulation, paving the way for more accurate, efficient, and less invasive procedures in various fields of microsurgery and scientific research.

\end{abstract}


\section{Introduction}

\begin{figure}[h]
\centering
\includegraphics[width=0.55\textwidth]{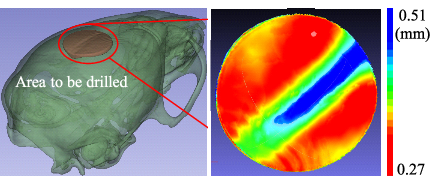}
\caption{The thickness distribution on the area to be drilled, which is thin and not uniform.} \label{fig:thickness}
\end{figure}

Biological specimens, unlike industrial components, exhibit significant variations in size and shape, and certain tissues possess a highly soft and deformable nature. This variability poses a challenge to robotic systems, which excel at performing repetitive or predefined motions, but struggle to automatically adapt to these variations. Consequently, the manipulation of biological specimens by robots often requires human operation, for instance, via a leader-follower robotic system. The complexity of this robotic manipulation increases when performed under a microscope, as in fields such as ophthalmology and neurosurgery where microsurgical tasks require high precision and accuracy \cite{Sha2019,vanMulken2018,Clarke2018,Takebe2019}. 
Similar technical challenges are encountered in the micromanipulation of cells, organoids, and small animals for medical research purposes. As a case in point, creating a skull window in mice, a procedure essential for medical science studies, presents significant technical challenges due to the need for microscopic manipulation \cite{Marinho2024}. The variability in skull thickness, ranging from 0.27 mm to 0.51 mm as shown in Figure \ref{fig:thickness} (based on \cite{Maga2017}), and the lack of a method to visualize the thickness distribution, hinders the accuracy and success rates of manual micromanipulation, leading to the exploration of robotic approaches. These approaches aim to improve precision while compensating for physiological tremors, as demonstrated by systems such as SmartArm for vitreoretinal surgery and Symani\footnote{https://www.mmimicro.com/our-technology/symani-surgical-system/} and MUSA-3\footnote{https://microsure.nl/musa} for plastic and reconstructive surgery \cite{Koyama2021,Aitzetmuller2022}. However, integrating preoperative information into these systems to further improve performance remains an area of development, particularly due to limitations in surface registration accuracy, which is essential for micromanipulation. 

The current study focuses on addressing these challenges in the mouse cranial window creation task, where an accurate skull thickness measurement from preoperative models is critical to prevent brain damage during drilling. In addition, accurate registration with the actual skull incorporating preoperative knowledge is essential for safe autonomous robotic micromanipulation. Although several registration methods have been developed using 2D/3D features such as bregma/lambda landmarks or surface curvature \cite{Spyrantis2022,Wu2021,Ghanbari2019}, their efficacy has primarily been demonstrated on bare or exposed skulls. However, practical application faces obstacles like limited skull surface exposure and obstruction of key features by biological materials.

\begin{table}[h]
\tbl{Existing registration methods for mice skulls}
{\begin{tabular}{lccc}
\hline
Methods       & \cite{Spyrantis2022} & \cite{Wu2021} & \cite{Ghanbari2019} \\ \hline
Target        & Bare Skull                           & Bare Skull             & Exposed Skull                       \\
Automation    & Manual                               & Autonomous             & Semi-Auto                                   \\
Perception    & Frame-based                 & \multirow{2}{*}{Depth Camera}                & A low-force contact \\
Approach      & Stereotaxy                &            & sensor         \\
Translational & \multirow{2}{*}{0.53 ± 0.22}          & \multirow{2}{*}{N/A}    & \multirow{2}{*}{0.011}        \\
Errors   (mm) &                                      &                          &                                             \\
Rotational    & \multirow{4}{*}{N/A}                 & $\left( \phi _X,\phi _Y,\phi _Z   \right) =$ & \multirow{4}{*}{N/A}   \\
Errors        &                                      & (0.17 ± 0.10,            &                                \\
(degree)      &                                       & 0.10 ± 0.04,            &                                \\
              &                                      & 0.14 ± 0.06)              &                               \\ \hline
\end{tabular}}
\label{table:1}
\end{table}

To improve registration accuracy, the use of 3D imaging technologies, including RGB-D systems that provide both 2D features and surface curvature, is essential. Previous studies, such as that of \cite{Probst2017}, have demonstrated the potential of stereomicroscope-based 3D vision systems in microsurgery, achieving a Root Mean Square Error (RMSE) of approximately 0.15 mm. Despite their accuracy, structured light-based systems are limited by processing speed \cite{Ly2021,Wang2022,Wu2021}, highlighting the challenge of balancing precision with operational efficiency in robotic manipulation. In response to these challenges, this study investigates advanced imaging and registration techniques to improve robotic control in micromanipulations in medical applications, aiming to address the need for both accuracy and efficiency in real-time operations.

Autonomous registration is necessary to plan motion for individual patients or biomedical specimens, and in particular, accurate micromanipulation in microsurgery customized for each patient will greatly contribute to his or her quality of life after surgery. It should also be mentioned that since the COVID-19 pandemic, the importance of the safety of medical professionals and scientists has been reaffirmed, and the robotization of medical procedures and medical research experiments has received renewed attention. Accurate registration technology is an important element for human-robot interaction, improving the accuracy of micromanipulation in medicine, improving the postoperative quality of life of patients, and protecting medical professionals in telerobotic medicine applications.

In this paper, we aim to develop a MSCS for robotic micromanipulation and registration for mice skulls with cranial features on a microscopic scale. This paper's contributions are threefold: (1) We have developed a MSCS employing a linear model that boasts an accuracy of 0.10 mm ± 0.02 mm and a processing speed of 30 frames per second (FPS), accelerated by a GPU-based disparity calculation algorithm; (2) We have demonstrated an accurate and precise registration scheme for the partially exposed cranial surface, utilizing CNN-based constrained and colorized registration methods; (3) We have integrated these methods and evaluated them through a mouse cadaver experiment.

\section{Materials and Methods}
\subsection{Requirements and Specifications}\label{sec:requirement} 

\begin{figure}[h]
\centering
\includegraphics[width=0.45\textwidth]{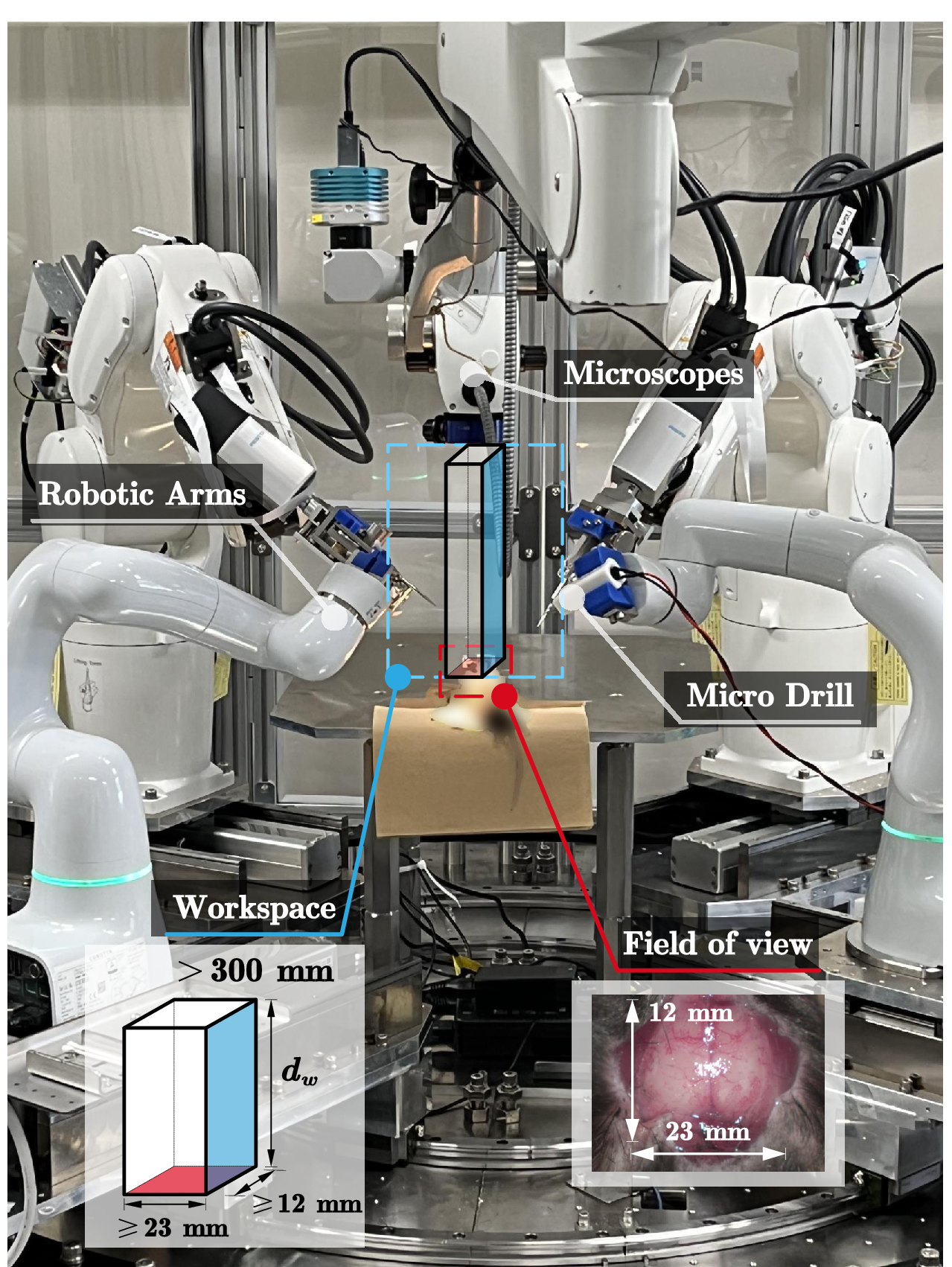}
\caption{The robotic system for microscopic manipulation \cite{Marinho2024} and the requirements for the 3D vision. $d_e$ is the effective working distance of the stereo microscope.}\label{fig:requirement}
\end{figure}

In the robotic platform for scientific experiments targeted in this work, the robotic end-effectors are spatially arranged for optimal functionality (see Figure \ref{fig:requirement} and Table \ref{table:2}). The field of view (FoV) is adjusted to accommodate a mouse skull specimen under these operation conditions. Regarding the required accuracy, from the template model proposed by the paper \cite{Maga2017}, it is found that the cranial skull's thickness ranges from 0.27 mm to 0.70 mm. The suture regions exhibit greater thickness (0.32 mm to 0.70 mm), while other continuous areas are thinner (0.27 mm to 0.43 mm). To accurately observe the thinning of the cranial skull, the system's depth resolution should be 0.14 mm or finer, which is approximately half of the minimum thickness of the cranial region. Furthermore, a frame rate exceeding 10 FPS is crucial to enable instant depth perception for human operators \cite{miller_response_1968}.

\begin{table}[h]
\tbl{Requirements for the 3D Vision System}
{\begin{tabular}{cccc}
\hline
Working Distance (mm)                       & FoV (mm)                             & Accu.(mm)       & FPS                 \\ \hline
300.00\textless{}d\textless{}600.00 & W\textgreater{}12.00, H\textgreater{}23.00 & \textless{}0.14 & \textgreater{}10 \\ \hline
\end{tabular}}
\tabnote{W and H are Width and Height of the FoV; Accu.: Accuracy.}
\label{table:2}
\end{table}

The FoV and the working distance requirements can be met by choosing a proper camera and macro lens setup. For a single microscope, follow the thin lens equation:
\[
\label{equ:dw}
d_w=f(1+\frac{1}{m})
\]
where $f$ is the focal length, $m$ is the magnification ratio, $d_w$ is the working distance of a single microscope.

According to the requirements, we adopted the hardware configuration in Figure \ref{fig:framework}, where the camera STC-HD853HDMI (Omron Inc., Japan) and the macro lens VS-LDA75 (VS Technology Inc., Japan) are used to capture microscopic images with a FoV 73.85$\times$41.54 $mm^2$ and $d_w$ around 500.00 mm ($f$=75 mm, m=0.175). A video capture card (DeckLink Quad HDMI Recorder, Blackmagic Design Inc., USA) was used in the capturing. The primary reason for choosing VS-LAD75 was its optimal FoV and $d_w$.

\subsection{A Framework for Microscopic Registration with MSCS}
\begin{figure}[ht]
\centering
\includegraphics[width=1\textwidth]{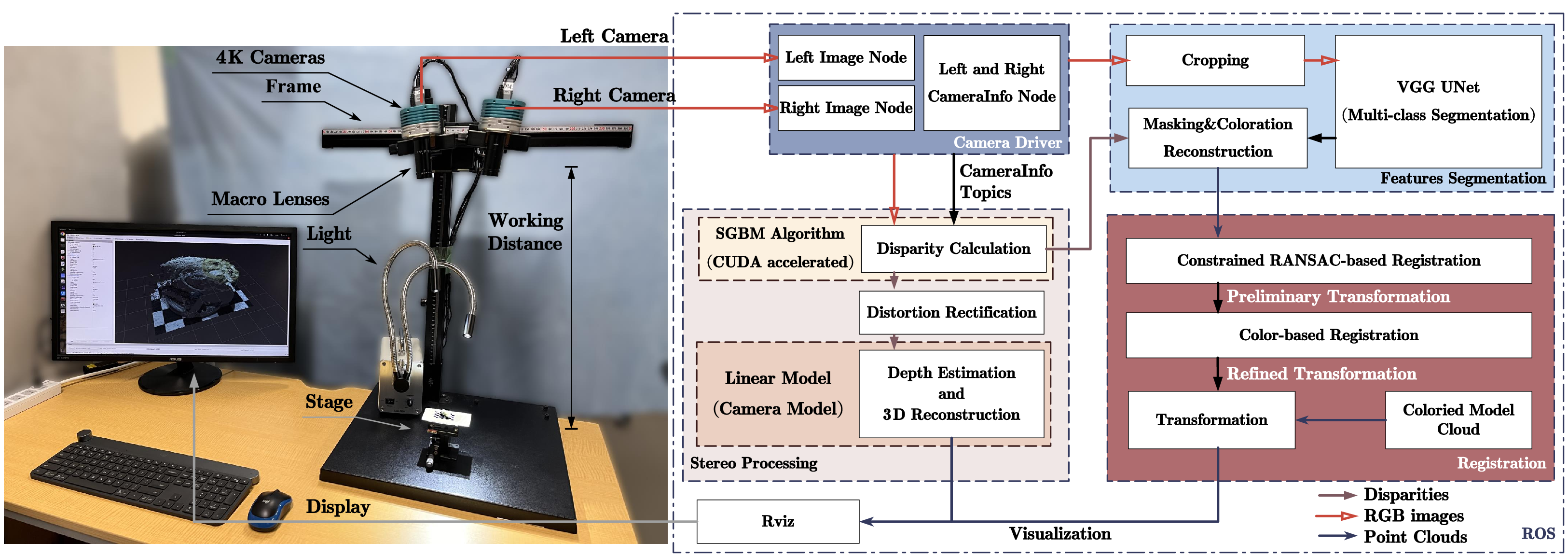}
\caption{A registration framework for microscopic manipulation, where the left side is the overview of the hardware of the MSCS, and the right side is the developed software in ROS. The diagrams on the right side show the workflow of the registration with each module.} \label{fig:framework}
\end{figure}

We developed a MSCS and a registration scheme in Robot Operating System (ROS), as shown in Figure \ref{fig:framework}. This framework addresses the challenges associated with microscopic 3D reconstruction and registration using cranial features. 

The specifications of hardware were discussed in Section \ref{sec:requirement} to meet the requirements. The MSCS captures stereo images from microscopes through the Camera Driver module. In the Stereo Processing module (Section \ref{sec:stereo}), the images are used to calculate disparity maps by semi-global block matching (SGBM) algorithm and reconstruct RGB point clouds by the linear model (Section \ref{sec:model}). The RGB-D images are utilized for feature segmentation and point cloud registration (Section \ref{sec:registration}). Finally, the reconstructed RGB point clouds and transformed model clouds are visualized in Rviz\footnote{http://wiki.ros.org/rviz}. 

\subsection{Stereo Processing Workflow in MSCS}\label{sec:stereo}

\subsubsection{Depth Perception with An Orthographic Projection Model}\label{sec:model}

\begin{figure}[h]
\centering
\includegraphics[width=0.45\textwidth]{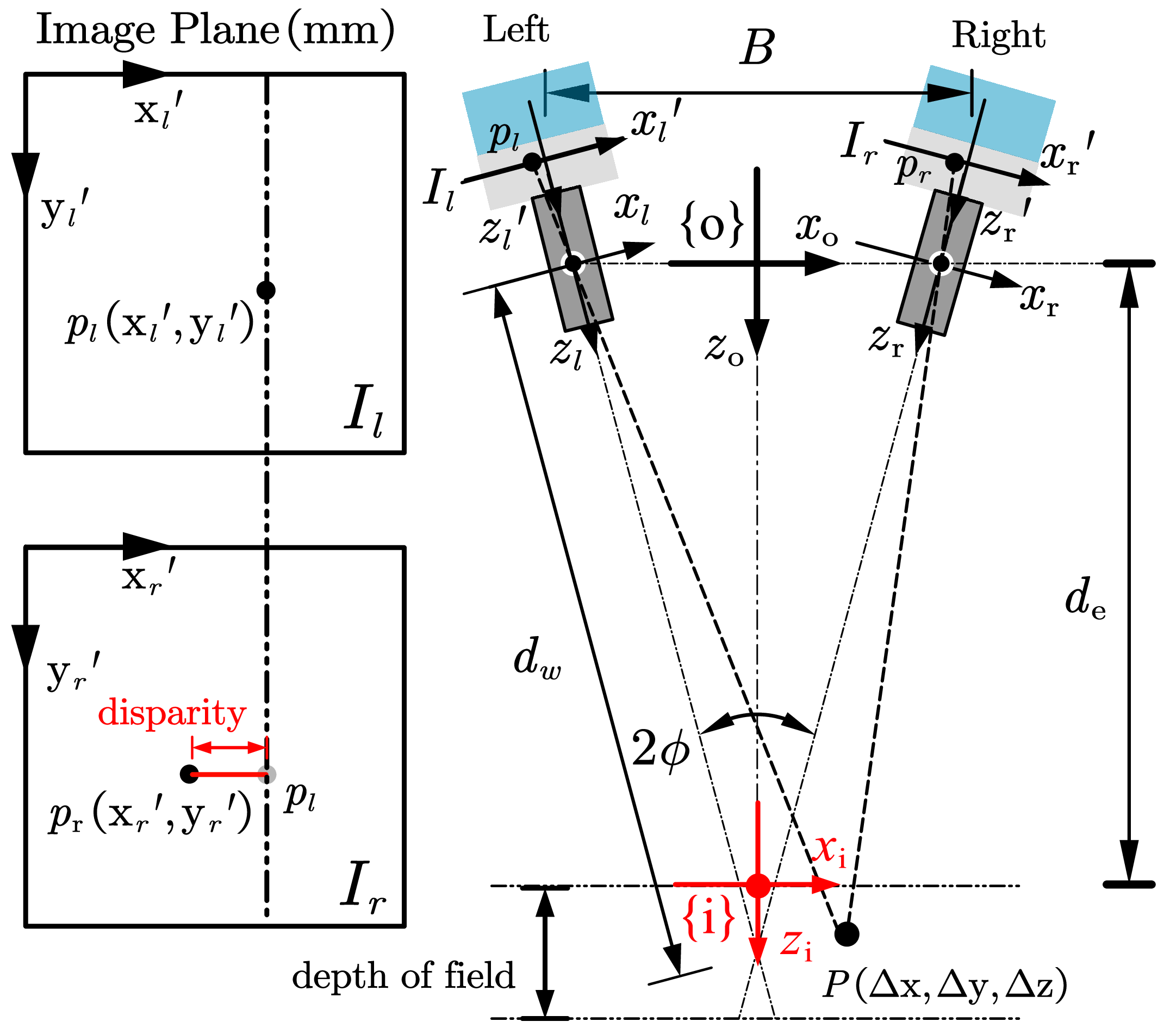}
\caption{Illustration of the stereo microscopy configuration. B is the baseline distance between the two cameras. The disparity is the distance between the projected points $p_l(x_l',y_l')$ and $p_r(x_r', y_r')$ in the image planes. The depth of field characterizes the setup's focal range. Frame $\{i\}$ is defined at an effective distance $d_e$ from the origin of Frame $\{o\}$ along the z-axis. $d_e$ is defined at the beginning of the depth of field of the stereo microscope. }\label{fig:model} 
\end{figure}

The hardware system layout depicted in Figure \ref{fig:framework} is conceptually simplified in Figure \ref{fig:model}, where we posit a point $P=(x_o, y_o,z_o)$ within a 3D coordinate system $\{o\}$. This point is simultaneously captured by both the left and right microscopes. The presentations of the point on the left and right microscopes are ($x_l,y_l,z_l$) and ($x_r,y_r,z_r$) regarding the microscopes' perspective centers, respectively. The distances from the point to the centers are denoted as $z_l$ and $z_r$. The projections of this point onto the image planes $I_l,I_r$ correspond to the pixels $p_l(x_l',y_l')$ and $p_r(x_r', y_r')$, respectively. We assume that the orientations of both microscopes are aligned within the same plane ($z_o-x_o$ plane) and converge at a fixed point, marked by a convergence angle of $2\phi$. 

The Pinhole Model is extensively applied for calibration and depth perception in stereo camera systems, addressing perspective effects as discussed by \cite{Kim1990}. The projection functions of the point on the image planes are defined based on the imaging geometry. For example, the projection in the left image plane $I_l$ can be expressed as:
\[
\label{equ:perspective}
\left( \begin{array}{l}
	x_l'\\
	y_l'\\
\end{array} \right) =-\frac{f}{z_l}\left( \begin{array}{l}
	x_l\\
	y_l\\
\end{array} \right) =\frac{-f}{(x_o+B/2)\sin \phi +z_o\cos \phi}\cdot \left( \begin{array}{c}
	(x_o+B/2)\cos \phi -z_o\sin \phi\\
	y_o\\
\end{array} \right) 
\]
where $f$ is the focal length of the macro lens, $z_l = (x_o+B/2)\sin \phi +z_o\cos \phi$ is deduced by coordinate rotation from Frame $\{o\}$ to left image plane frame $\{l\}$. 

$z_o$ in the denominator term equals point's z-position $\Delta z$ in Frame $\{i\}$ adding an effective distance $d_e$. Thus, the denominator term $z_l$ can be expressed regarding Frame $\{i\}$:
\[
\label{equ:zo}
z_l = (x_o+B/2)\sin \phi +z_o\cos \phi = (\Delta x + B/2)\sin \phi +(d_e + \Delta z)\cos \phi
\]

In the context of stereo microscopy, where the FoV ($\Delta x_{max}, \Delta y_{max}$) and the depth of field ($\Delta z_{max}$) are significantly smaller than the working distance $d_e$, we specifically address the calculation of $z_l$ under the configuration $d_e \approx d_w$ = 500.00 mm, $\phi$ = $7^o$, B $\approx$ 135.00 mm, $|\Delta x_{max}|$ = 41.54 mm, $|\Delta z_{max}| \approx$ 10.00 mm: 
\[
\label{equ:perspective2}
d_{e} \cos \phi+(B / 2) \sin \phi \approx 503  \gg |\Delta x_{max}| \sin \phi+|\Delta z_{max}| \cos \phi\approx 13
\]

Consequently, the disparity-depth mapping function approximates an orthographic projection as per \cite{Kim1990} for $0 \ll (z-d_w) \ll |\Delta z_{max}|$: 
\begin{align*}
\label{equ:o_projection}
z 
& =\frac{1}{2 \sin \phi}\left(\frac{h}{m}+B \cos \phi\right)\\
\left(\begin{array}{l} x_{l}' \\ y_{l}' \end{array}\right)
& =m\left(\begin{array}{c}(x + B/2) \cos \phi - z \sin \phi \\ y \end{array}\right)
\end{align*}
where the magnification ratio $m=-f/z_o$. 

This projection presupposes alignment of the epipolar plane with the horizontal image coordinate axes of both microscopes' views, ensuring distortion-free orthographic projection through the use of low-distortion macro lenses.

For practical calibration convenience using a checkerboard, we further streamline to a linear function (the linear model):
\begin{equation}
\label{equ:lm}
\begin{array}{l}
z=d_e+\Delta z=d_e+h/h_{\rho}\\
x=\Delta x=(x_l'-c_x)P_{\rho x}\\
y=\Delta y=(y_l'-c_y) P_{\rho y}\\
\end{array}
\end{equation}
A point ($x,y,z$) in microscopic 3D space is projected to the left image of the microscope at a pixel ($x_l',y_l'$) (pixel), where $h_\rho $ (pixel/mm) denotes the depth response to unit disparity in reality, $h$ (pixel) is the pixel's disparity value, $d_e$ (mm) is the effective working distance of the stereo microscope setup (indicating the physical distance from Frame $\{i\}$ to Frame $\{o\}$), $c_x,c_y$ (pixel) are the camera's principal point image coordinates. $P_{\rho x},P_{\rho y}$ (pixel/mm) indicate the object's pixel sizes in the x and y directions, respectively. The corresponding pixel $p_r(x_r', y_r')$ (pixel) in the right image is essential for calculating the disparity value using a CUDA-accelerated SGBM algorithm \cite{Noauthor2023}.

\subsubsection{Model Calibration and Distortion Rectification}\label{sec:distortion}

In the context of the linear model, we calibrated factors such as $h_\rho ,P_{\rho x},P_{\rho y}$ using checkerboard images. For instance, we estimated the checkerboard's pose (Square size: 0.50 mm x 0.50 mm) and identified the corners. The corners' pixel distances and their true lengths (0.50 mm) relative to the pose were computed and averaged to determine the pixel size in millimeters, yielding values for $P_{\rho x},P_{\rho y}$. For $h_\rho $, step-height experiments were conducted by incrementally elevating a z-axis stage and observing the corresponding disparity responses, from which the factor's value was deduced by fitting a linear curve.

\begin{figure}[h]
\centering
\includegraphics[width=0.55\textwidth]{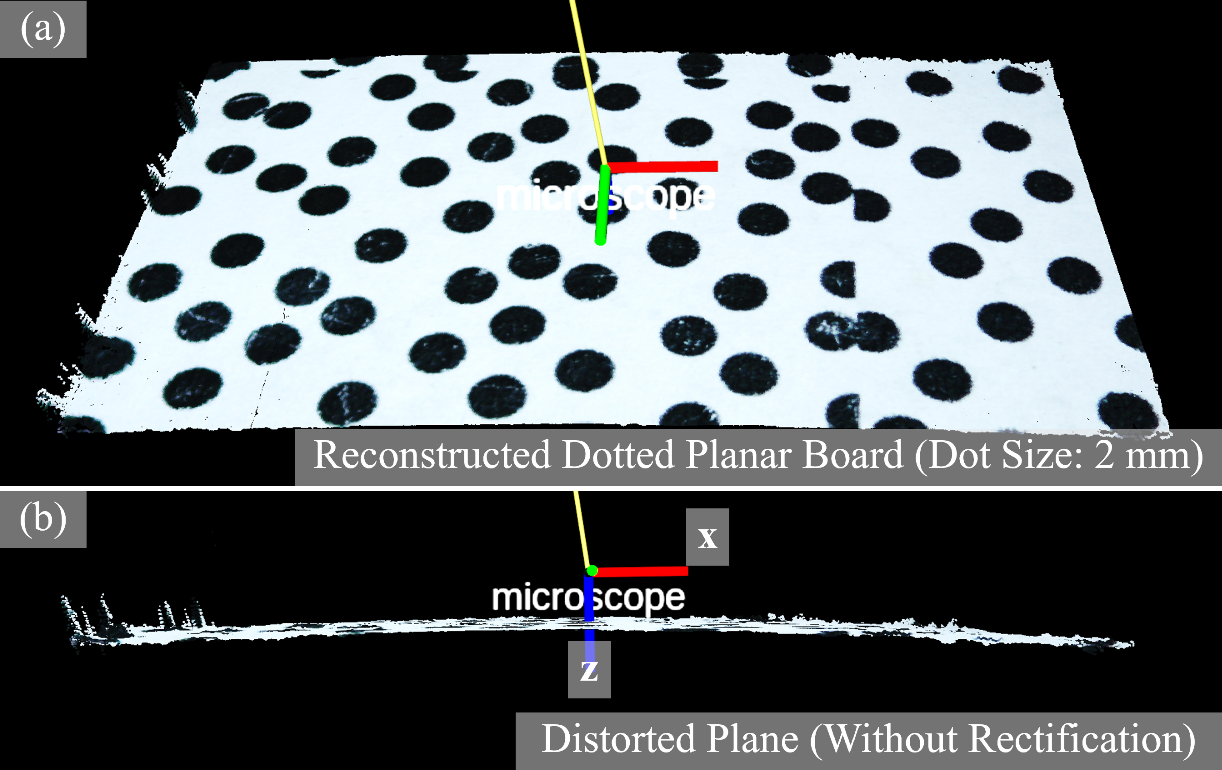}
\caption{A distorted planar board reconstruction by the stereo microscope, which is similar to CMO distortion.}\label{fig:distortion} 
\end{figure}

As per \cite{Danuser1999}, a stereo microscope utilizing a common main objective (CMO) introduces specific CMO distortion. Although our research's setup does not incorporate a CMO, the reconstruction of a planar object revealed a similar distortion mode as illustrated in Figure \ref{fig:distortion}.

Given that the distortion-induced errors are uniformly distributed along the vertical axis, as identified by \cite{Danuser1999}, we adopted a similar assumption for our analysis. To counteract these distortions, we differentiated between a horizontal planar object (illustrated by the dotted pattern in Figure \ref{fig:distortion}(b)) and its z-axis transition from an ideal to a distorted planar surface. This method allowed us to effectively compensate for the constant residuals introduced by the distortion, ensuring a more accurate representation of the observed object.

\subsection{The Registration Scheme with Pre-operative Planning}\label{sec:registration}

\subsubsection{Pre-operative Planning and Cranial Features for Registration}

\begin{figure}[h]
\centering
\includegraphics[width=0.55\textwidth]{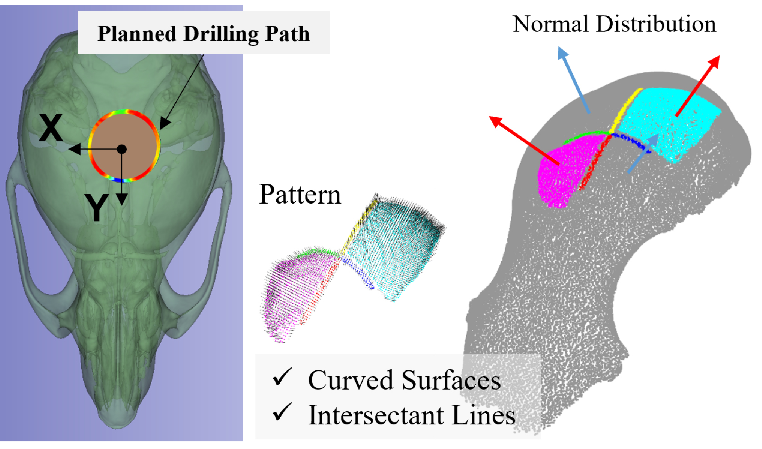}
\caption{Left: Demonstrates a synthesized template of a laboratory mouse's cranium \cite{Maga2017} alongside a pre-planned circular patch designated for removal; Right: Highlights cranial features that are pivotal for the drilling task, featuring curved surfaces for orientation guidance and intersectant lines for enhanced positional accuracy.}\label{fig:preplanning} 
\end{figure}

For pre-operative planning utilizing CT-scan imagery, the selection of laboratory mice as subjects requires the utilization of micro-computed tomography (micro-CT) to precisely capture the craniofacial phenotype in three dimensions as shown in \ref{fig:preplanning}. The segmentation editor tool within the 3D Slicer\footnote{https://www.slicer.org/} software facilitates the delineation of the targeted circular area (marked in a color gradient), tailored to the precise location and thickness distribution.

Adjacent to the drilling site, certain features facilitate the model's alignment with the intraoperative RGB point cloud. From the colorless reconstructed model in Figure \ref{fig:preplanning}, only sutures and surfaces, discernible through the crevices, provide essential spatial information for successful registration. Curved surfaces convey critical orientation information, while intersectant lines (the sutures) pinpoint the exact spatial positioning necessary for pattern recognition.

\subsubsection{Proposal of The Registration Scheme}

\begin{figure}[ht]
\centering
\includegraphics[width=1\textwidth]{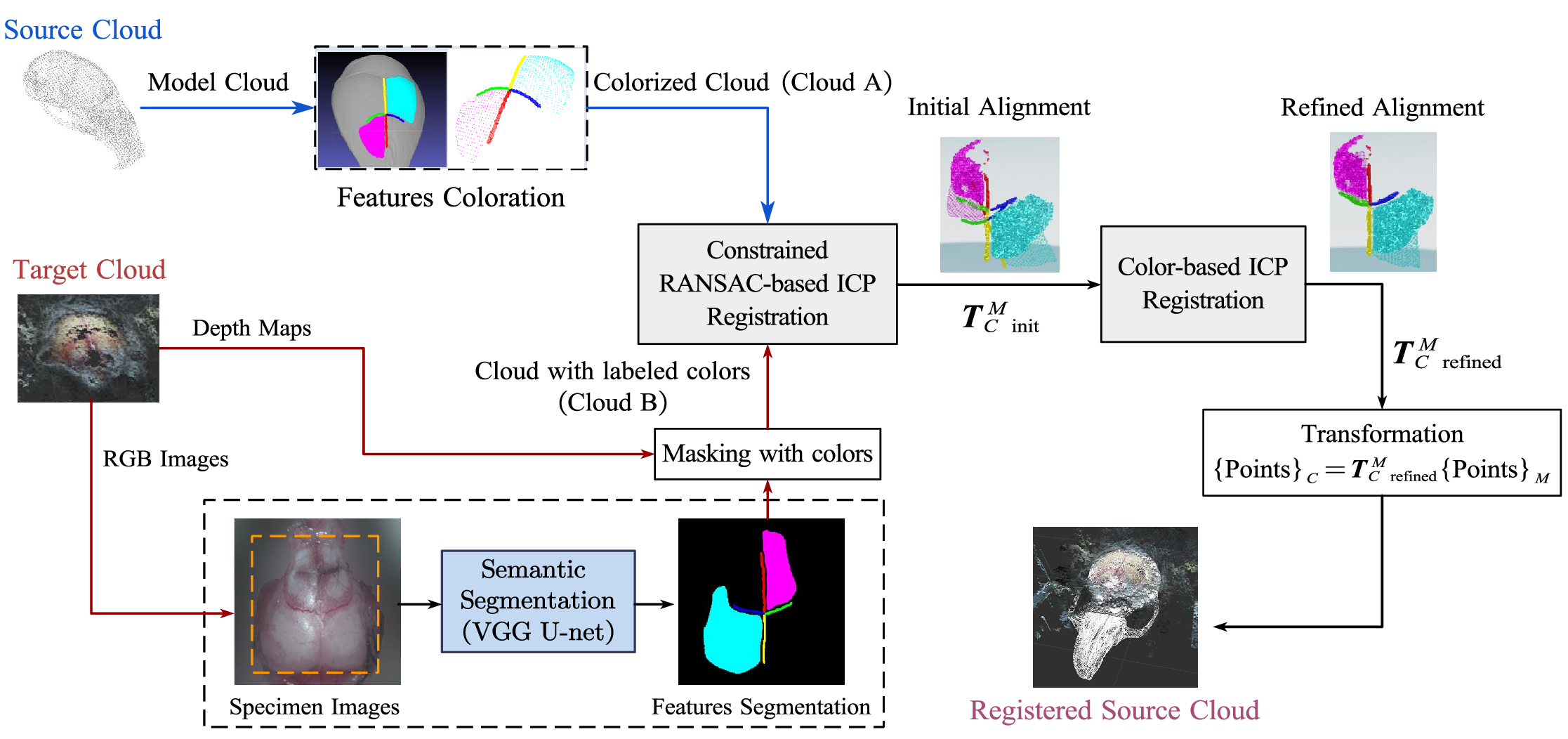}
\caption{A registration scheme for patterned features in a colorless model, where the features should be recognizable both in the model and the reconstructed RGB point cloud. The random sample consensus (RANSAC) technique is used for outlier filtering. The iterative closest point (ICP) method is widely used in point cloud registration.}\label{fig:scheme} 
\end{figure}

Our registration scheme matches patterned features of a colorless model with those in the corresponding RGB point cloud via manual coloration and multi-class feature segmentation, as outlined in Figure \ref{fig:scheme}. 

\paragraph{Features Coloration in Model} Initially, features in the model including intersectant lines and curved surfaces are colorized with labeled colors based on distinct RGB value discrimination (e.g., [1,0,1], [0,1,1]), as depicted in Figure \ref{fig:scheme}. Employing varied colors corresponding to different orientations or feature locations aids the color-based ICP algorithm \cite{Park2017} in executing precise matching. 

\begin{figure}[h]
\centering
\includegraphics[width=1\textwidth]{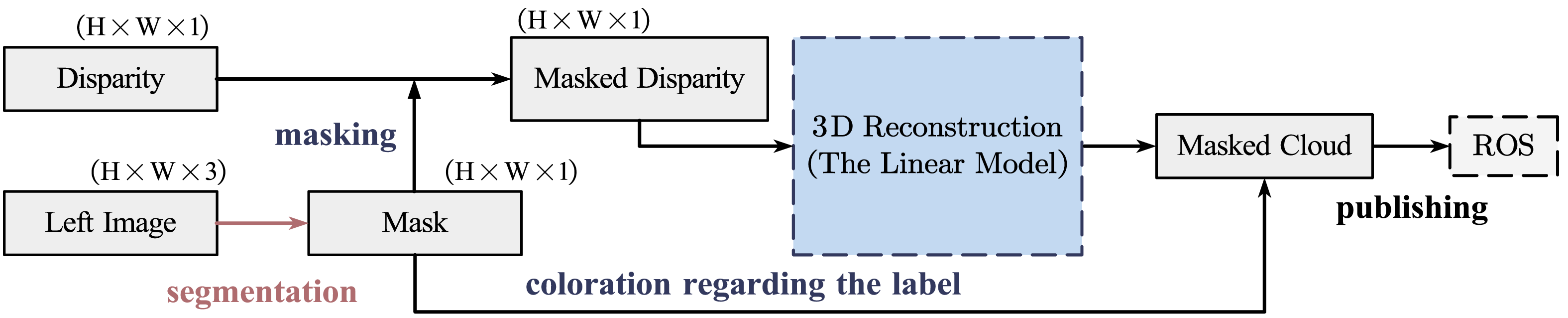}
\caption{Masked-cloud generation and coloration method by the segmentation results}\label{fig:masking} 
\end{figure}

\paragraph{Features Segmentation from RGB-D images} Underlying this scheme is the premise that cranial features are identifiable from RGB imagery of mice skulls. Thus, the features can be segmented with a multi-class semantic segmentation algorithm, where in this research, we used a VGG U-net model \cite{Gupta2023}, as demonstrated in Figure \ref{fig:scheme}. For mouse skull registration, a minimalist feature set is preserved, including four sutures and two facial areas between sutures, selected for their distinctive orientation information. The features are segmented into different classes through the neural network. The masks with class labels are used to segment the depth map as illustrated in the workflow in Figure \ref{fig:masking}. The resultant class-labeled masks facilitate depth map segmentation, as depicted in the workflow of Figure \ref{fig:masking}. Consequently, point clouds pertaining to features with designated RGB colors are color-matched rather than their original colors.

\paragraph{Point Cloud Registration} The process begins with aligning the preoperative model's colorized point cloud (Cloud A) with the feature-segmented point cloud with labeled colors (Cloud B) to achieve registration within the microscopic 3D space. Given potential noise from reconstruction or segmentation misclassification in Cloud B, an initial alignment utilizing the RANSAC-based ICP method filters outliers and aligns the point clouds with the estimated surface normal of the surface features. The registration is constrained by its 2D orientation concerning the left image frame. The iteration of the color-based ICP method \cite{Park2017} is initialized with the output transformation ${\mathbf{T}^M_{C}}_{\mathrm{init}}$. This method chiefly refines alignment by exploiting the colorized inline feature clouds. The finalized outcome is a polished transformation matrix  ${\mathbf{T}^M_{C}}_{\mathrm{refined}}$ employed to adjust the preoperative model.

\section{Experimental results}

\subsection{Measuring Experiments}

\subsubsection{Evidence of the Linear Relationship in Depth Estimating}

\begin{figure}[h]
\centering
\includegraphics[width=0.55\textwidth]{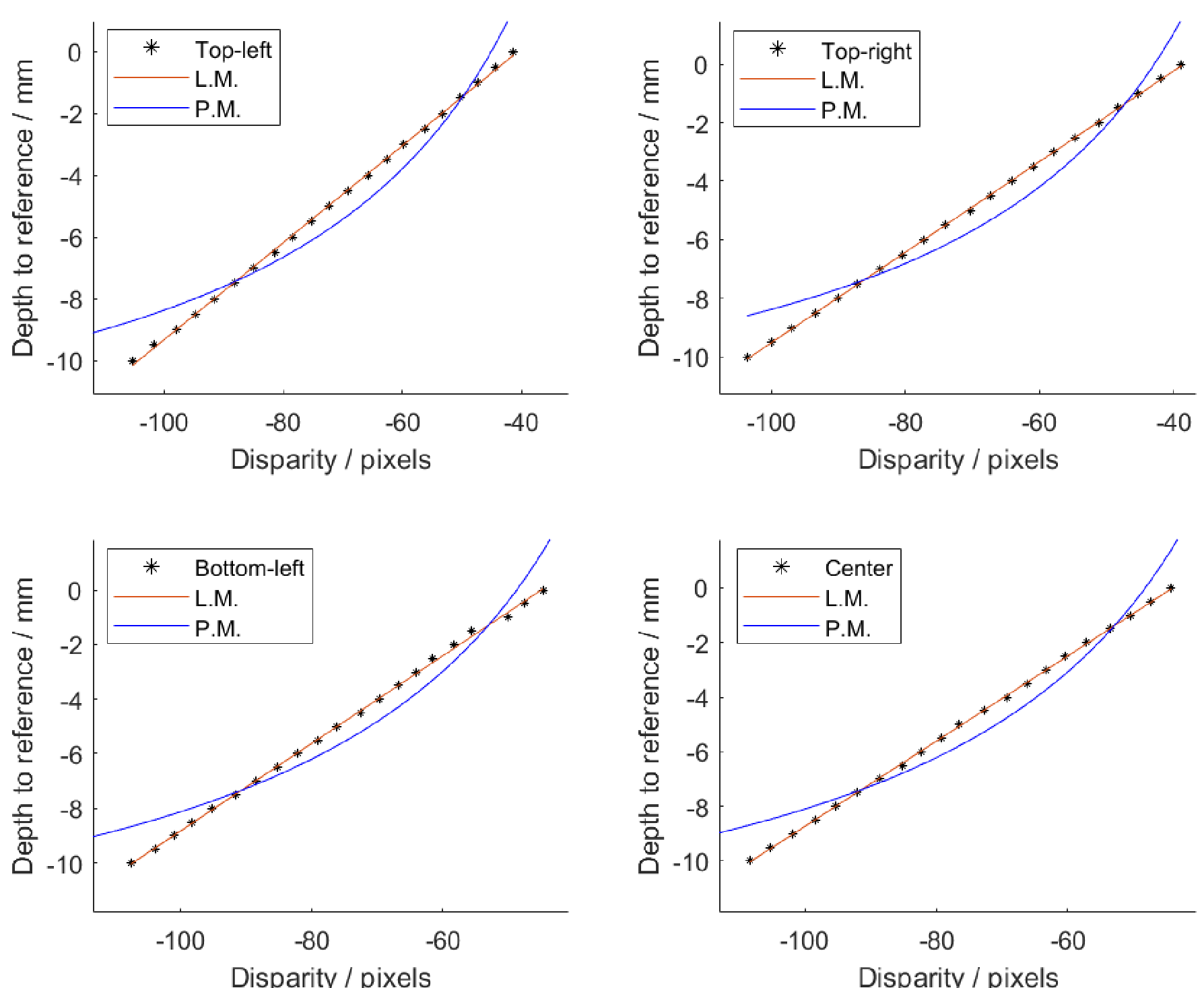}
\caption{Step height measuring experiments: four positions (top-left, top-right, bottom-left, center) are sampled to measure the disparity-depth response. The black points are the measurements. The red line is the fitness result of the Linear Model (L.M.); The blue line is the result of the Pin-hole Model (P.M.).}\label{fig:step_height} 
\end{figure}

To ascertain whether the disparity-depth mapping function aligns more closely with a linear function than the conventional pinhole camera model, we conducted a comparative experiment. This involved a step-height measurement experiment within the field of view, focusing on the relative depth response to disparity changes. We incrementally elevated the stage along its z-axis by 0.50 mm increments, recording disparity values ten times at each position to obtain an average. Given a depth of field of approximately 10.00 mm, we documented 21 steps, as illustrated in Figure \ref{fig:step_height}.

The experimental outcomes indicate a linear response (Figure \ref{fig:step_height}), with L.M. achieving a fit characterized by ($R^2$, RMSE, MAE)=(0.9996, 0.06, 0.05), in contrast to P.M., which yielded ($R^2$, RMSE, MAE)=(0.9418, 0.73, 0.64). Here, $R^2$ represents the coefficient of determination, RMSE, and MAE the mean absolute error, demonstrating the linear model's superior accuracy and consistency in depth estimation.

\subsubsection{Experiments for Evaluating the Depth Resolution}

\begin{figure}[h]
\centering
\includegraphics[width=0.55\textwidth]{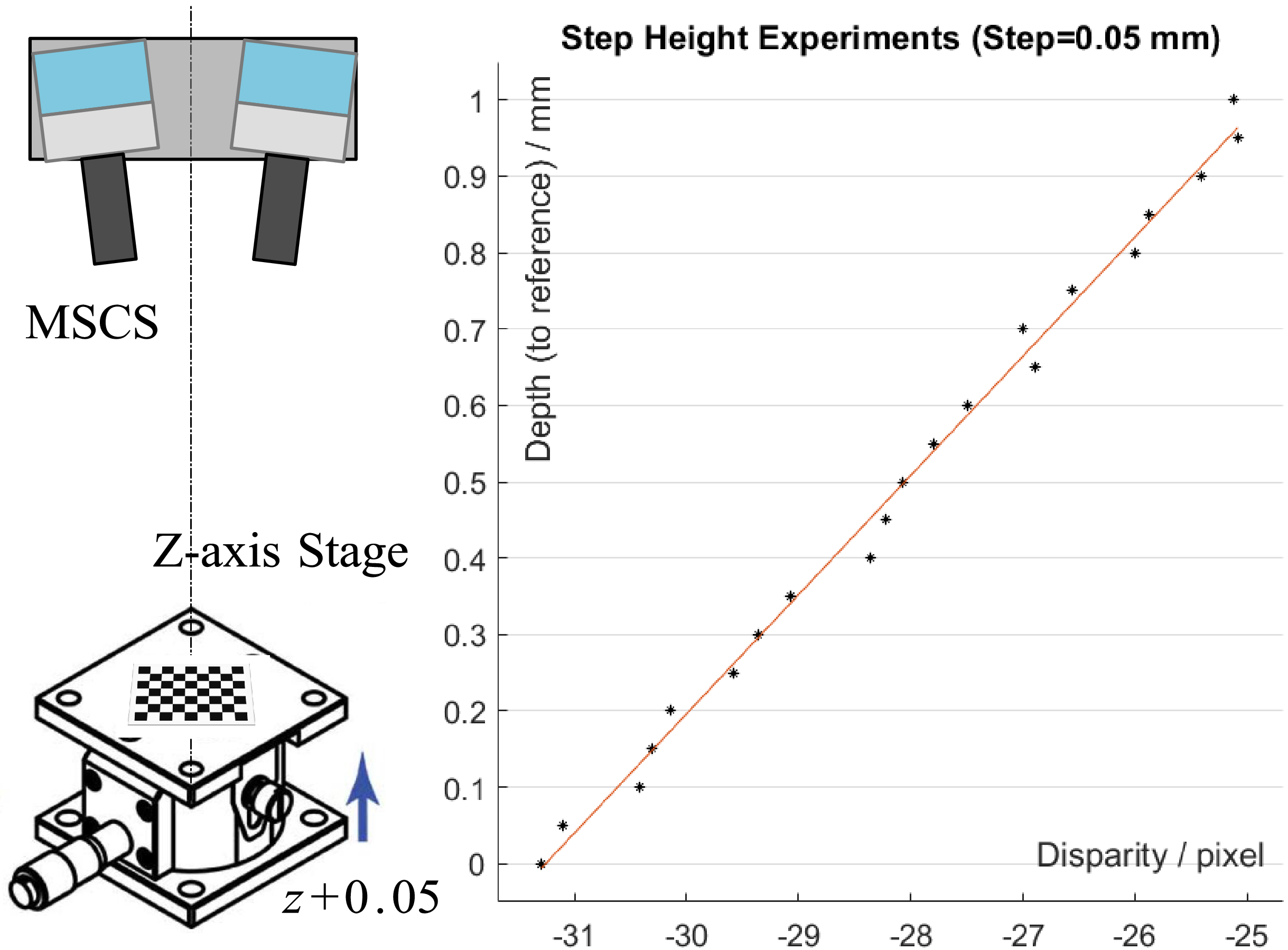}
\caption{Step height experiments from 0.00 to 1.00 mm with a step 0.05 mm}\label{fig:resolution} 
\end{figure}

A subsequent step-height experiment was designed to gauge the depth perception accuracy facilitated by the linear model, as depicted in Figure \ref{fig:resolution}. This experiment involved adjusting the z-axis stage in 0.05 mm increments, with disparity measurements taken at each step, spanning a total of 21 samples from 0.00 mm to 1.00 mm. The results revealed a linear fit with $R^2$=0.9941 and an RMSE of 0.02 mm at a resolution of (540$\times$960), signifying an enhancement over the 3D reconstruction error reported by \cite{Probst2017} (approximately 0.15 mm at 1080P). Furthermore, the RMSE of 0.02 mm indicates that the average measurement error falls below the minimum detectable distance difference (0.10 mm), affirming the system's capability to discern minute changes of 0.05 mm with uniform distribution. Consequently, we deduced that our system possesses a resolution of 0.10 mm in depth estimation.

\subsection{Surface Registration Experiments}

\subsubsection{Experimental Objective and Setup} 
\begin{figure}[h]
\centering
\includegraphics[width=0.55\textwidth]{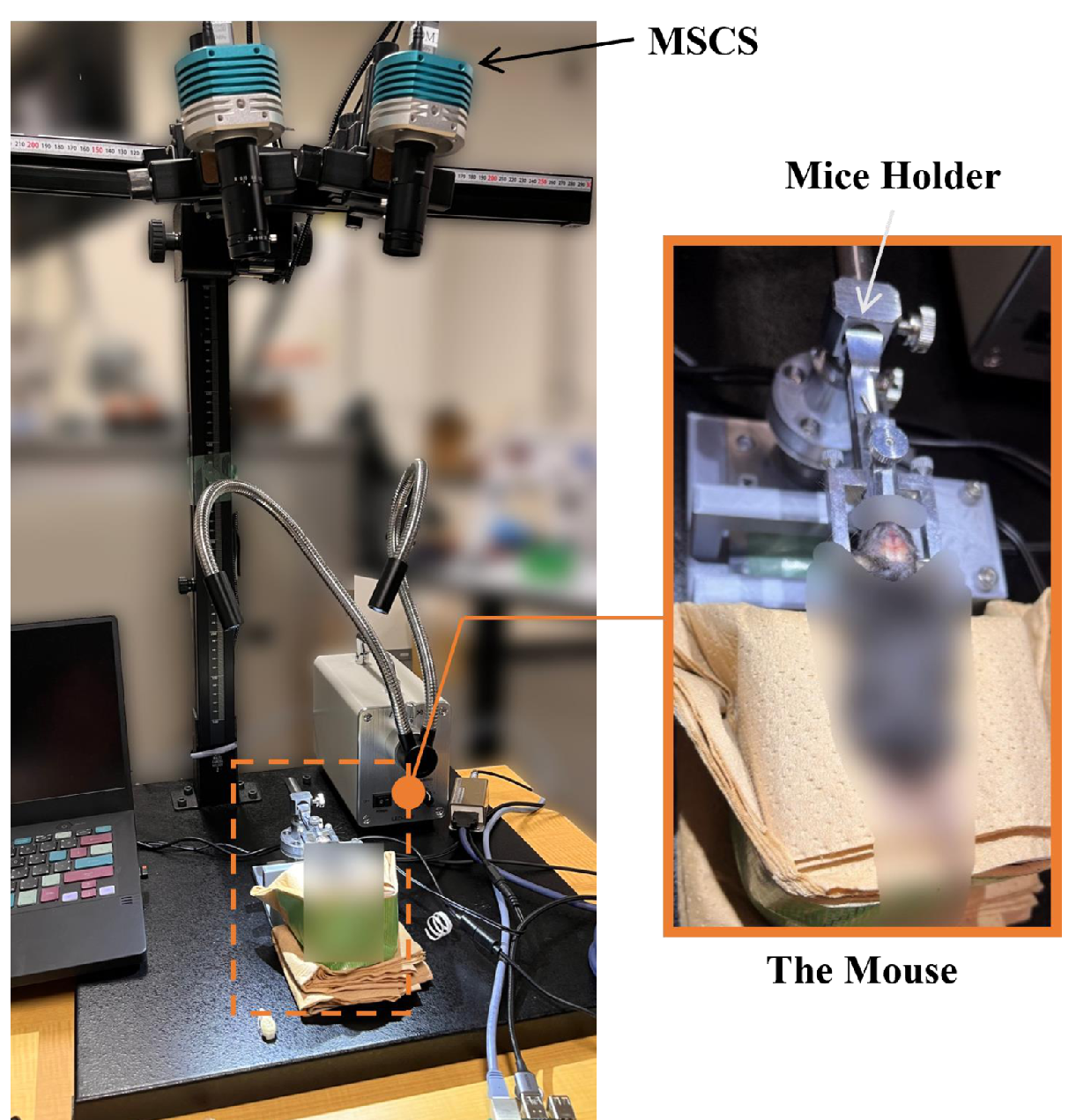}
\caption{Experimental setup, a euthanized mouse fixed to a holder was used for the experiments.}\label{fig:setup} 
\end{figure}

To evaluate the precision of the registration scheme, an euthanized 16-week-old mouse was used for the reconstruction and registration (refer to Figure \ref{fig:setup}). It is crucial to note that the mouse, provided by an adjacent research laboratory, was not specifically euthanized for this study but would have otherwise been disposed of. 

We collected a training dataset (156 pictures, 512x512) from 10 different mice (including the dataset for testing the scheme, which has 10 of the 156 images) to train the network for the segmentation task. After augmentation, we fed the augmented dataset (9516 pictures, 512x512) to train the network, where the number of classes is 7, including 6 feature classes and 1 undefined class. The training was completed using a workstation with a Ryzen 9 5950X (AMD Inc., USA) and an RTX3080 (Nvidia Inc., USA). 

The efficacy of the proposed registration scheme was evaluated through practical experiments involving a colorized model, as shown in Figure \ref{fig:preplanning}.  As the training of the semantic segmentation network involved the images for validating the registration scheme, the results show the capability of the scheme under the condition of correct segmentation.

Leveraging the ROS-based system development enabled the capture of the cranial surface reconstruction sequence, utilizing the rosbag\footnote{http://wiki.ros.org/rosbag} package for real-time replay within ROS. 

Quantitative assessment of the registration's precision entailed analyzing a specific sequence consisting of 105 frames, captured while the mouse head was static and fixed by a mouse head holder (SG-4N, Narishige, Japan). Although the reconstruction occurred in real-time, the mouse skull's position was stationary, ensuring consistency in the evaluation of registration precision. 

\subsubsection{Evaluation Criteria and Results} 

\begin{figure}[h]
\centering
\includegraphics[width=1\textwidth]{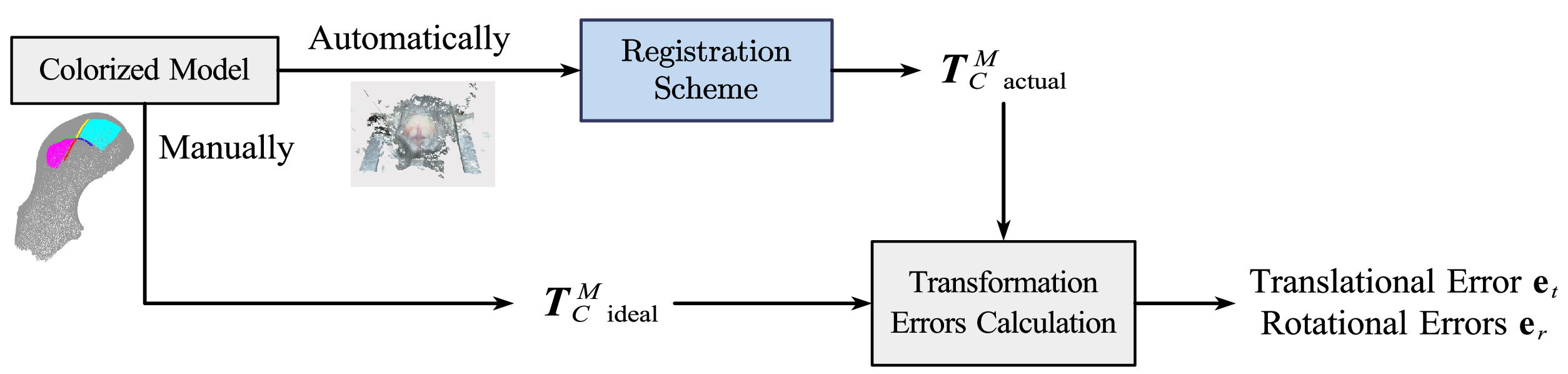}
\caption{Quantitative evaluation method through manual alignment.}\label{fig:evaluation} 
\end{figure}

Due to the absence of a reference frame within the reconstructed mouse point clouds, we manually aligned the mouse bregma model with the cranial surface, utilizing the same feature chosen in the coloration. The translational and rotational errors were then assessed through a workflow illustrated in Figure \ref{fig:evaluation}. 

\begin{figure}[h]
\centering
\includegraphics[width=1\textwidth]{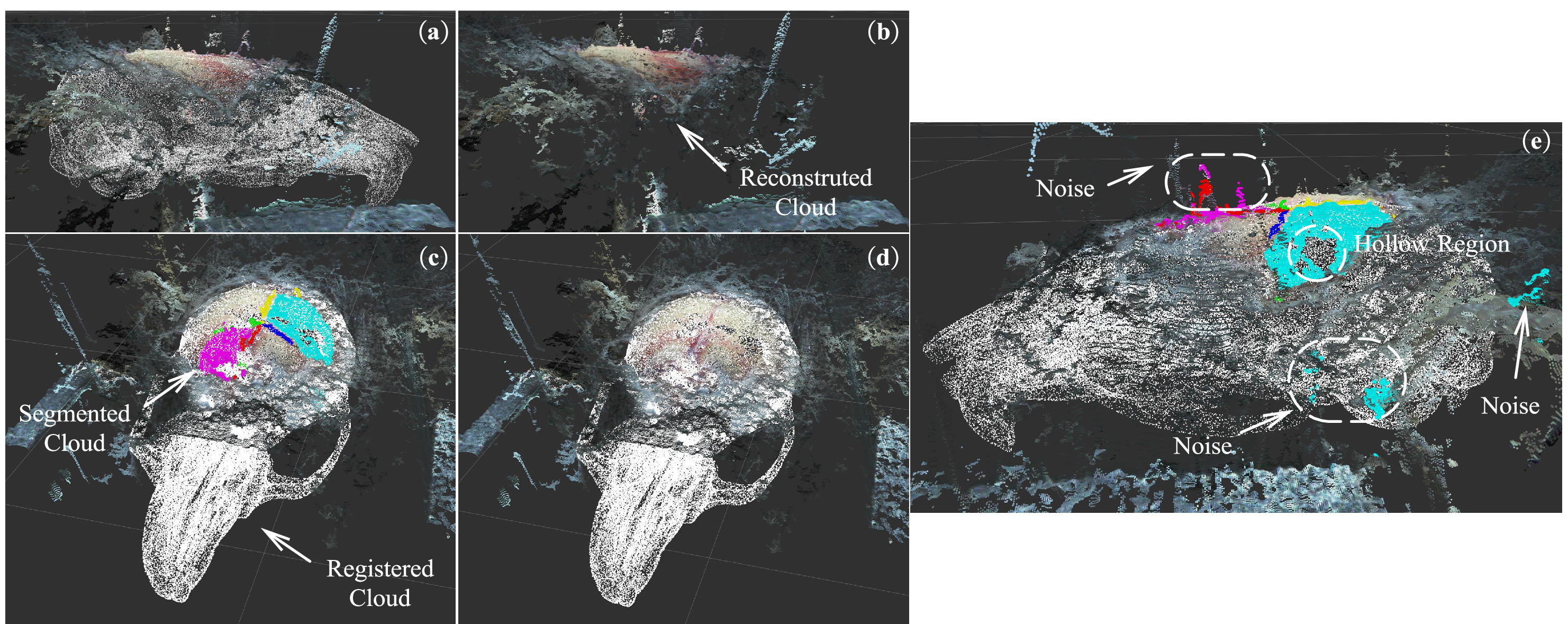}
\caption{One of the registration results, where the white clouds are cloud transformed by autonomous registration, and the colorful clouds (Cloud B) are segmented and colorized points.}\label{fig:vis} 
\end{figure}

Following the autonomous registration process, the transformed bregma cloud was rendered within a 3D vision space, depicted as a white point cloud in Figure \ref{fig:vis}. Despite the presence of noise in Cloud B, as evidenced in Figure \ref{fig:vis}(e), the registration scheme demonstrated resilience to such disturbances, achieving accurate alignment with the cranial surface. This noise could originate from light reflections that failed to match any features during disparity calculations or from segmentation misclassifications.

\begin{figure}[h]
\centering
\includegraphics[width=0.55\textwidth]{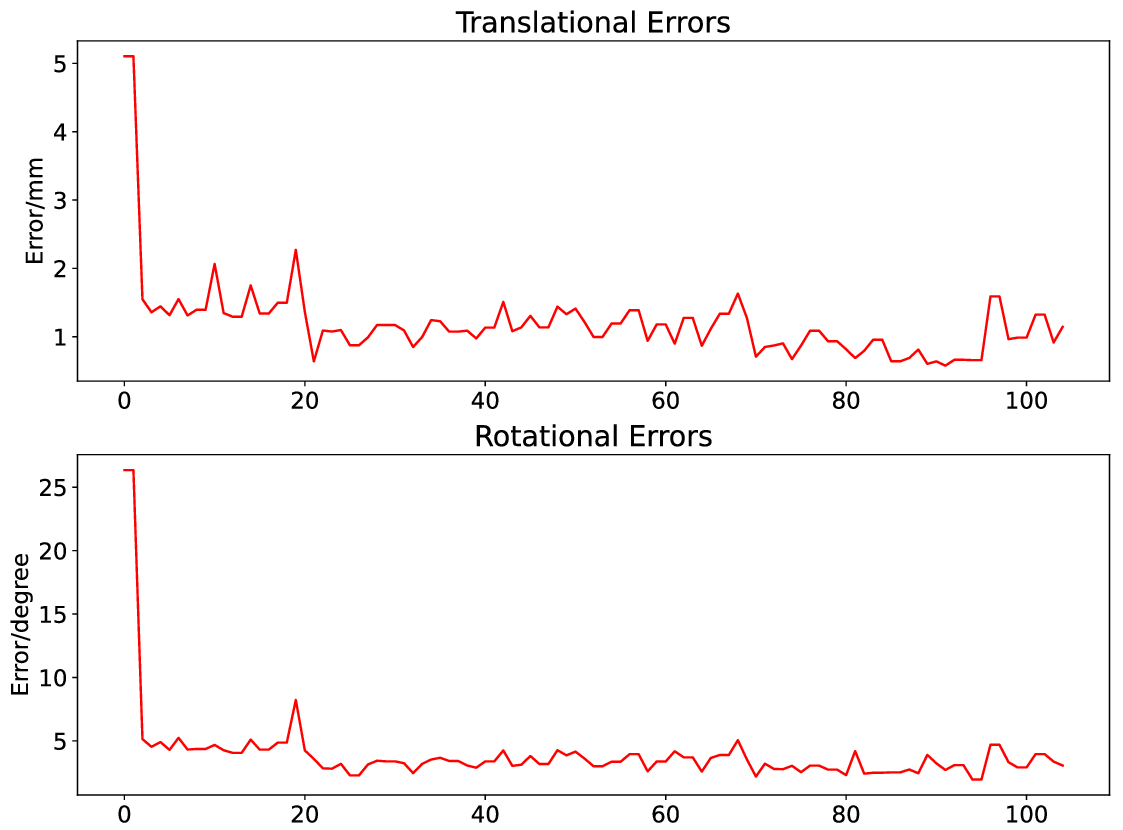}
\caption{Registration errors evaluated by 105 independent and continuous RGB-D frames.}\label{fig:errors1} 
\end{figure}

Error calculations between the registered and manually aligned frames were conducted using the following equations:
\begin{align*}
\mathbf{e}_t&=\left| \mathbf{t}_1-\mathbf{t}_2 \right| _2 \\
\mathbf{e}_r&=\mathrm{arccos} \left( \frac{\mathrm{Tr}\left( \mathbf{R}_1\mathbf{R}_{2}^{T} \right) -1}{2} \right)\
\end{align*}
Here, $\mathrm{Tr}\left( \mathbf{R}_1\mathbf{R}_{2}^{T} \right)$ represents the trace of $\mathbf{R}_1\mathbf{R}_{2}^{T}$, with $\mathbf{e}_t$ and $\mathbf{e}_r$ denoted in millimeters and degrees, respectively. Figure \ref{fig:errors1} delineates the translational and rotational errors across a sequence of 105 frames, highlighting an average processing duration of approximately 1.60 seconds for the entire scheme according to the nodes' publication frequency in ROS. These findings underscore the system's stability and robustness in registering static objects amidst variations in lighting and occlusions.

\begin{figure}[h]
\centering
\includegraphics[width=0.55\textwidth]{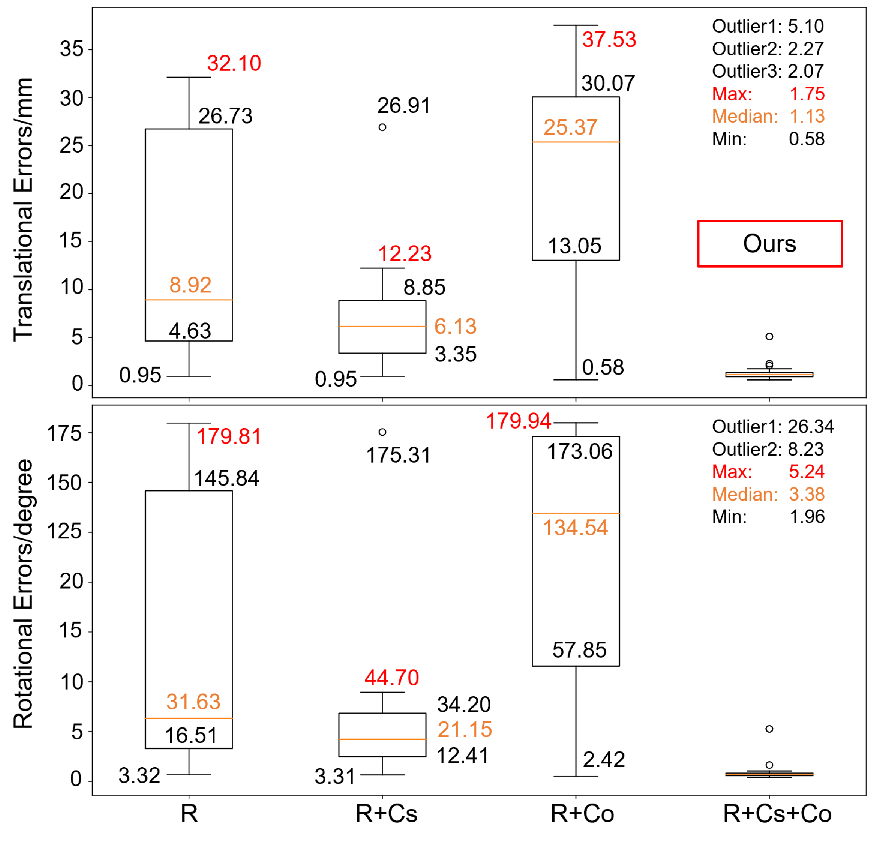}
\caption{Registration errors compared to other combinations, where R is the RANSAC technique, Cs is the constraints and Co is the colors.}\label{fig:errors2} 
\end{figure}

A comparative analysis with other methods, employing the identical sequence, is presented in Figure \ref{fig:errors2}. This box plot contrasts four different combinations of registration methods, delineating variations with and without constraints (Cs) or colorization (Co). While the minimum errors across the methods were comparably similar, notable differences were observed in their deviations. The application of constraining techniques effectively limited deviations, yet lacked precision with random distributions. Color-based registration yielded the lowest error rates among the tested methods, although it demonstrated susceptibility to noise within the point cloud. A synergistic integration of both techniques culminated in the most favorable outcomes—minimizing deviation and registration errors. Excluding outliers shown in Figure \ref{fig:errors2}, the scheme attained the lowest average translational error (mean 1.13 ± 0.31 mm, excluding outliers 5.10, 2.27, and 2.07 mm) and rotational error (mean 3.38° ± 0.89°, excluding outliers 26.34° and 8.23°). As the outliers mainly occurred at the beginning of the registration, the errors show the registration accuracy when the transformation became stable.

\section{Discussion}

This study introduced the integration of a Microscopic Stereo Camera System (MSCS) augmented by a simplified stereo microscopy model, termed the linear model, for depth perception. Demonstrating high precision in the step height experiment, the system attained 0.10 mm accuracy with an RMSE of 0.02 mm, as detailed in Figure \ref{fig:resolution}. Meeting the real-time performance criteria outlined in Section \ref{sec:requirement}, the system's processing capability has been optimized to achieve up to 30 FPS in ROS, making it suitable for both manual and automated interventions. This optimization enables it to outperform comparative methods (\cite{Ghanbari2019,Ly2021,Wang2022,Probst2017,Wu2021}), ensuring efficient and effective operation as listed in Table \ref{table:3}. Nonetheless, the system's performance on biological specimens is hindered by reflective surfaces, which compromise disparity calculations due to lost correlations. The proposed linear model can provide accurate relative depth reconstruction. Although we defined the effective working distance $d_e$ for its use in stereo microscopy, achieving precise calibration for absolute depth reconstruction remains a challenge. For practical implementation, the effective working distance can be set to 0 without compromising registration accuracy. Further endeavors will aim to improve the system's robustness to the reflection and accurate calibrating of the working distance. 

\begin{table}[h]
\tbl{Comparison of depth perception methods in related work}
{\begin{tabular}{@{}lcccccc@{}}
\toprule
Works                          & \cite{Ghanbari2019} & \cite{Ly2021} & \cite{Wang2022} & \cite{Probst2017} & \cite{Wu2021} & Ours                  \\ \midrule
\multirow{2}{*}{Methods}       & \multirow{2}{*}{Force}               & \multirow{2}{*}{SL}            & \multirow{2}{*}{SL}              & Dense                          & \multirow{2}{*}{SL}            & \multirow{2}{*}{SGBM} \\
                               &                                      &                                &                                  & disparity                      &                                &                       \\
$d_e$ (mm)                     & N/A                                  & N/A                            & 89$\sim$96                       & N/A                            & 250                            & 500                   \\
FoV (mm)                       & N/A                                  & 30×35                          & 3.9°-0.6°$^1$                       & 12×23                          & 135×222                        & 22×35                 \\
\multirow{2}{*}{Accuracy ($\mu$m)} & \multirow{2}{*}{11}                  & \multirow{2}{*}{98.5 ± 4.5}    & ± 4.0 (X,   Y)                   & $\sim$150                      & \multirow{2}{*}{10 ± 2}        & 100 ± 20 at           \\
                               &                                      &                                & ± 7.5 (Z)                        & at 1080P                       &                                & 540×960               \\
FPS                            & N/A                                  & 0.01                           & N/A                              & 5                              & 0.9                            & 30                    \\ \bottomrule
\end{tabular}}
\tabnote{\textsuperscript{1}with a range of view 6.0-1.0 mm; SL: Structured Light.}
\label{table:3}
\end{table}

Further, a novel registration scheme was developed to leverage this system for microscopic registration on patterned surfaces. The challenges in the registration include 1) little 3D features on the exposed cranial surface, 2) partial visibility or occlusion by the biological tissues or the instrument, and 3) inconsistent feature spaces between the preoperative model with channels (x,y,z) and the intraoperative point cloud with channels (x,y,z,R,G,B). According to \cite{Zhao-regis-2023}, existing state-of-the-art registration strategies can be categorized into two types: End-to-End and hybrid methods. The End-to-End method directly learns features from the point clouds through the neural network and regresses the transformation of the source clouds. However, applying this kind of method is difficult mainly because of the inconsistent feature spaces between the model and the intraoperative cloud. Therefore, the research focuses on developing a hybrid method for registration. Current hybrid methods detect key points from the point clouds for matching. Regarding the biological specimens, key point detection is not robust enough as the operation has unpredictable occlusions. Besides, the anatomical landmarks may be hard to recognize because the appearance of each individual is different. Inspired by contextualization in \cite{Unberath2021}, which extracts semantic information from 2D or 3D data to handle the inconsistencies between the feature spaces in medical image registration, we investigated related research in this field to register two data with inconsistent feature spaces. However, another strategy using contour features for registration like \cite{Koo2022} is not applicable to the task, as there are no fixed contour features on the partially exposed cranial surface.

Therefore, regarding the limitations of current registration strategies with hybrid methodology, we developed a registration scheme with a novel feature descriptor for the registration between the preoperative model and the intraoperative RGB point cloud of a biological specimen on a microscopic scale in Section \ref{sec:registration}. Rather than landmarks or contours, the scheme selects the lines and surface features for registration, where, in a mouse case, they are the sutures and surfaces on the cranial surface. The feature includes enough position and orientation information for accurate registration. Another novel method is converting the contexts to labeled colors to utilize the context for registration. Instead of keeping the original RGB information, we segmented the features semantically from RGB point clouds as multiple classes and labeled colors regarding the classes. In the preoperative model, we painted the points belonging to the same context with the same context color. Registration is performed between the two clouds using the same context colors.

We compared the feature descriptor with others in Table \ref{tab:regis_strategy_comparisons}. The proposed method can handle more complicated features and achieve more accurate registration than others, showing the proposed method's advancement.

\begin{table}
\caption{Comparison of the state-of-the-art registration strategies}
\label{tab:regis_strategy_comparisons}
\begin{adjustbox}{max width=1.2\textwidth,center}
    \begin{tabular}{lllll}
    \hline
                                       & End-to-End                                  & \multicolumn{3}{l}{Hybrid   Strategy (Contextualization)}                                                                                \\ \hline
    Feature descriptor                 & Neural Network                              & Key-points-based          & Contour-based                               & {Surface + Intersectant lines} \\
    Feature space                      & Cloud features                              & Points                    & Lines                                       & Points/Lines/Surfaces                                          \\
    Channel                            & XYZ+(RGB)                                   & XYZ                       & XYZ                                         & XYZ+RGB (labeled)                                              \\ \hline
    \multicolumn{5}{l}{Requirements}                                                                                                                                                                                            \\ \hline
    Inconsistent channels              &-                                           &+ &+                   &+                                      \\
    Partial visibility                 &                     &                           &                     &                                        \\
    (Occlusion)                        & \multirow{-2}{*}{+} & \multirow{-2}{*}{-}       & \multirow{-2}{*}{+} & \multirow{-2}{*}{+}                    \\
    Surface Curvature                  &                     &                           &                                             &                                        \\
    (Orientation)                      & \multirow{-2}{*}{+} & \multirow{-2}{*}{-}       & \multirow{-2}{*}{-}                         & \multirow{-2}{*}{+}                    \\
    Individual Differences$^1$ &-                                           &-                         &+                   &+                                      \\
    Dataset                & Inaccurate                                  & Inaccurate                & Inaccurate                                  & Accurate                               \\ \hline
    \end{tabular}
    \end{adjustbox}
\tabnote{$^1$Appearances change regarding different individuals.}
\end{table}

The MSCS and the registration scheme presented have broad applications, extending to microsurgeries and micro-electromechanical system (MEMS) manufacturing, as discussed in the introduction. The registration scheme is particularly applicable in scenarios like craniotomy or plastic and reconstructive surgery, where surfaces with preoperative models possess visible patterned features, despite being limited to microscopic views. 
In addition to reconstruction accuracy for such applications, it is imperative to recognize the importance of ensuring safety in robot-human interactions. Therefore, further efforts will also focus on applying the developed technologies to other applications to demonstrate how our technologies can contribute to robotic micromanipulation in medicine, for example, to ensure patient safety in microsurgery or to ensure the safety of medical researchers while successfully performing robotic micromanipulation tasks in a remote space.

\section{Conclusions}
 
This study develops and evaluates a Microscopic Stereo Camera System (MSCS) and a registration scheme, addressing the challenges in microscopic manipulation for scientific and surgical applications. The MSCS demonstrates remarkable accuracy (0.10 mm ± 0.02 mm) and real-time processing capability (30 FPS), laying a solid foundation for precise control in delicate procedures. Additionally, the registration scheme, employing CNN-based methods tailored for patterned surfaces with intersectant lines and normal features, exhibits rapid yet stable performance even in complex scenarios. Notably, in the application of mice skull registration, the translational error is measured at 1.13 mm ± 0.31 mm, with a rotational error of 3.38° ± 0.89° over 105 continuous frames, achieved within a 1.60 seconds per registration. In the future, the technologies developed will be integrated into other applications to ensure both the performance and safety of micromanipulation in surgery and medical research.

\section*{Acknowledgement(s)}

The authors would like to thank our research collaborators and scientists for their fruitful comments, especially Professor Murilo Marques Marinho, for his kind help in improving this work.

\section*{Disclosure statement}

All authors declare no financial or personal relationships with other people or organizations.

\section*{Ethical approval}

The animal had been euthanized in another study and was reused in our study, thus no ethical approval was required.

\section*{Funding}

This study was supported in part by JST Moonshot R\&D JPMJMS2033-09 and in part by the University of Tokyo Fellowship.

\bibliographystyle{tfnlm}
\bibliography{manuscript}

\end{document}